\documentclass[a4paper,11pt,twocolumn,twoside]{article}
\usepackage[dvips]{graphicx}
\usepackage{sepln_en}
\usepackage{fullname}
\usepackage[utf8]{inputenc}
\usepackage[english]{babel} 
\usepackage{color,soul}
\usepackage{url}
\usepackage[colorinlistoftodos]{todonotes} 

\title{Language Resources in Spanish for Automatic Text Simplification across Domains}

\author {\textbf{Antonio Moreno-Sandoval,$^1$} \textbf{Leonardo Campillos-Llanos,$^2$} \textbf{Ana García-Serrano$^3$}\\
$^1$Facultad de Filosofía y Letras, Universidad Autónoma de Madrid\\
$^2$ILLA - Consejo Superior de Investigaciones Científicas\\
$^3$Facultad de Informática, Universidad Nacional de Educación a Distancia\\
\texttt{antonio.msandoval@uam.es}, \texttt{leonardo.campillos@csic.es}, \texttt{agarcia@lsi.uned.es}
 \\
}

\seplntranstitle{ Recursos lingüísticos en español para simplificación automática de textos en varios dominios}

\seplnclave{Simplificación Automática de Textos, Recursos Lingüísticos, Corpus.}

\seplnresumen{En este artículo se describen los recursos lingüísticos y modelos de simplificación automática de textos en español desarrollados para tres dominios: financiero, médico y de estudios históricos. Se han creado varios corpus en cada dominio, guías de anotación y simplificación, un léxico de términos médicos técnicos y simplificados, conjuntos de datos empleados en competiciones de dominio financiero, y dos herramientas de simplificación. La metodología, los recursos y las publicaciones asociadas son públicas (\url{https://clara-nlp.uned.es/}).}

\seplnkey{Automatic Text Simplification, Language Resources, Corpora.}

\seplnabstract{This work describes the language resources and models developed for automatic simplification of Spanish texts in three domains: Finance, Medicine and History studies. We created several corpora in each domain, annotation and simplification guidelines, a lexicon of technical and simplified medical terms, datasets used in shared tasks for the financial domain, and two simplification tools. The methodology, resources and companion publications are shared publicly on the website: \url{https://clara-nlp.uned.es/}.} 

\firstpageno{1}

\begin{document}


\setlength\titlebox{15cm} 

\label{firstpage} \maketitle

\section{Introduction}

Simplifying administrative discourse \cite{cunha2022redactor}, financial documents or medical information is necessary for laymen citizens to participate actively in their daily decisions. Current international initiatives, exemplified by the International Plain Language Federation\footnote{\url{https://www.iplfederation.org/}} or the Plain Language Association International (PLAIN),\footnote{\url{https://www.plainlanguagenetwork.org/}} have addressed these demands and created guidelines for plain language writing \cite{escribirclaridad}. 

Automatic text simplification \cite{saggion2017automatic} can produce plain language versions of financial, historical or medical documents \cite{segura2017simplifying} and is a complement to the final human revision using dedicated tools \cite{espinosa2023review}. This task requires linguistic resources such as corpora \cite{alarcon2023easier}, benchmarks \cite{gonzalez-dios-etal-2022-irekialfes,madina2023easy}, guidelines or lexicons \cite{ferres2022alexsis}.

The CLARA-NLP coordinated project, funded by the Spanish Government, gathered three research teams from Universidad Autónoma de Madrid (UAM), Consejo Superior de Investigaciones Cientificas (CSIC) and Universidad Nacional de Educación a Distancia (UNED). The main goal is the exploration and integration of knowledge on text simplification, and the development of language resources and Deep Learning (DL) models for Spanish in the three domains. The three multidisciplinary teams incorporated NLP experts and specialists in each domain (economists, historians, physicians and specialists in medical communication). 

This article reports the produced resources for automatic simplification of Spanish texts (ASST) in three domains (Finance, Medicine and History studies). In the following, we describe the objectives (\S\ref{objectives}), the methods (\S\ref{methods}) and the results (\S\ref{results}): corpora, guidelines, lexicons, simplification tools and datasets for shared tasks.

\begin{table*}[htbp!]
\renewcommand{\arraystretch}{1.75}
\begin{center}
\scriptsize{
\begin{tabular}{|l|l|l|}
\hline \bf Domain & \bf Resource & \bf Description and availability \\ \hline
\parbox[t][][t]{1.5cm}{Digital \\ Humanities} & \parbox[t][][t]{3.15cm}{ Corpus } & \parbox[t][][t]{10.35cm}{CLARA-DM
corpus: 37 newspapers manually transcribed (201 pages), 143 newspapers automatically transcribed (657 pages), and ten newspapers manually annotated (53 pages; 24 843 tokens). The dataset is available at \url{https://github.com/CLARA-HD/diario-madrid-dataset}. The transcribed texts are available searching at \url{https://albali.lsi.uned.es/buscador
}\vspace{5pt} } \\ \cline{2-3}
& \parbox[t][][t]{3.15cm}{ Guideline } & \parbox[t][][t]{10.35cm}{NER annotation guideline (available on request). \vspace{5pt} } \\ \cline{2-3}
 & \parbox[t][][t]{3.15cm}{ Models } & \parbox[t][][t]{10.35cm}{The model for automatic transcription 
(Spanish print XVIII-XIX) is already available
for free use at the Transkribus site: \url{https://readcoop.eu/transkribus}. 
\\ NER models for CLARA-DM corpus, using the defined taxonomy (explained in subsection \ref{results}): monolingual RoBERTa, multilingual XLM-RoBERTa, RoBERTa-BNE-NER-CAPITEL, XLM-RoBERTa-NER-Spanish and XLM-RoBERTa-NER-HRL. These models are described on \cite{salidoetal2023}. \vspace{5pt} } \\ 
\hline
Financial & \parbox[t][][t]{3.15cm}{ Corpus } & \parbox[t][][t]{10.35cm}{ Spanish-English parallel corpus of annual reports, including 20 different companies; 896 978 words in Spanish, 778 328 words in English, and a total of 15 788 segments. \vspace{5pt}  } \\ \cline{2-3}
& \parbox[t][][t]{3.15cm}{ Guidelines } & \parbox[t][][t]{10.35cm}{Three annotation guidelines: one for the discourse marker tagger, another for the term and keyword extractor, and the guideline for the Spanish FinCausal dataset. \vspace{5pt} } \\ \cline{2-3}
& \parbox[t][][t]{3.15cm}{ Lexicon } & \parbox[t][][t]{10.35cm}{ 13 869 terms from a corpus of financial reports, validated by terminologists. \vspace{5pt} } \\ \cline{2-3}
& \parbox[t][][t]{3.15cm}{ Models } & \parbox[t][][t]{10.35cm}{ A Spanish Discourse Marker tagger and an Automatic Term Extractor based on BERT family models. \vspace{5pt} } \\ \cline{2-3}
& \parbox[t][][t]{3.15cm}{ Simplification system } & \parbox[t][][t]{10.35cm}{ SimpFin financial term
simplification tool: \url{http://leptis.lllf.uam.es/simfin} } \\ \cline{2-3}
& \parbox[t][][t]{3.15cm}{ Shared-tasks datasets } & \parbox[t][][t]{10.35cm}{Three datasets in Spanish for the Financial Narrative Processing workshops: FinTOC2022, FNSummarisation 2022 and 2023, and FinCausal 2023. \vspace{5pt} } \\  \hline
Medical & \parbox[t][][t]{3.15cm}{ Corpus } & \parbox[t][][t]{10.35cm}{ CLARA-MeD corpus: 24 298 pairs of professional and simplified texts (over 96 million tokens): \url{https://digital.csic.es/handle/10261/269887}. \\ 5000 parallel (technical/simplified) sentences: 3800 semi-automatically aligned and manually revised (149 862 tokens) + 1200 manually simplified by linguists (144 019 tokens): \url{https://digital.csic.es/handle/10261/346579} \\ Also on the Hugging Face hub: \url{https://huggingface.co/datasets/CLARA-MeD/} \vspace{0.1pt} } \\ \cline{2-3} 
& \parbox[t][][t]{3.15cm}{ Simplification guidelines } & \parbox[t][][t]{10.35cm}{ Syntactic and lexical simplification criteria, available at: \\ \url{https://digital.csic.es/handle/10261/346579} \vspace{5pt} } \\ \cline{2-3}
& \parbox[t][][t]{3.15cm}{ Lexicon } & \parbox[t][][t]{10.35cm}{14 013 pairs of technical/simplified terms, including gender/number variants and conjugated forms: \url{https://digital.csic.es/handle/10261/349662} \vspace{5pt} } \\ \cline{2-3} 
& \parbox[t][][t]{3.15cm}{ Models } & \parbox[t][][t]{10.35cm}{Simplification models (Multilingual BART, mT5, NASES, Pegasus XSUM and BERTIN Alpaca) on Hugging Face: \url{https://huggingface.co/CLARA-MeD} 
\\ Medical entity recognition models also available on the Hugging Face hub: \url{https://huggingface.co/medspaner}
\vspace{5pt} } \\ \cline{2-3} 
& \parbox[t][][t]{3.15cm}{ Simplification system } & Medical terms simplification tool: \url{http://claramed.csic.es/demo} \vspace{1pt} \\ \hline
\end{tabular}
}
\end{center}
\vspace*{-5mm} 
\caption{\label{resources} Resources released in the CLARA-NLP project.}
\end{table*}

\section{Developed Resources for ASST: Methods}
\label{objectives}
\label{methods}

We collected several resources for simplification-related tasks across domains. Table \ref{resources} lists the released resources for each domain. The first step was collecting representative corpora to annotate, simplify and conduct experiments (\S\ref{corpora}). Annotation and simplification guidelines were produced for several tasks (\S\ref{guidelines}). A medical lexicon was also created by extracting terminology from the medical corpus (\S\ref{lexicon}). In the last stages, we performed experiments with state-of-the-art Deep Learning models for simplification and named entity recognition (NER) (\S\ref{models}). 

\begin{figure}[htb!]
 \includegraphics[width=7.5cm,clip]{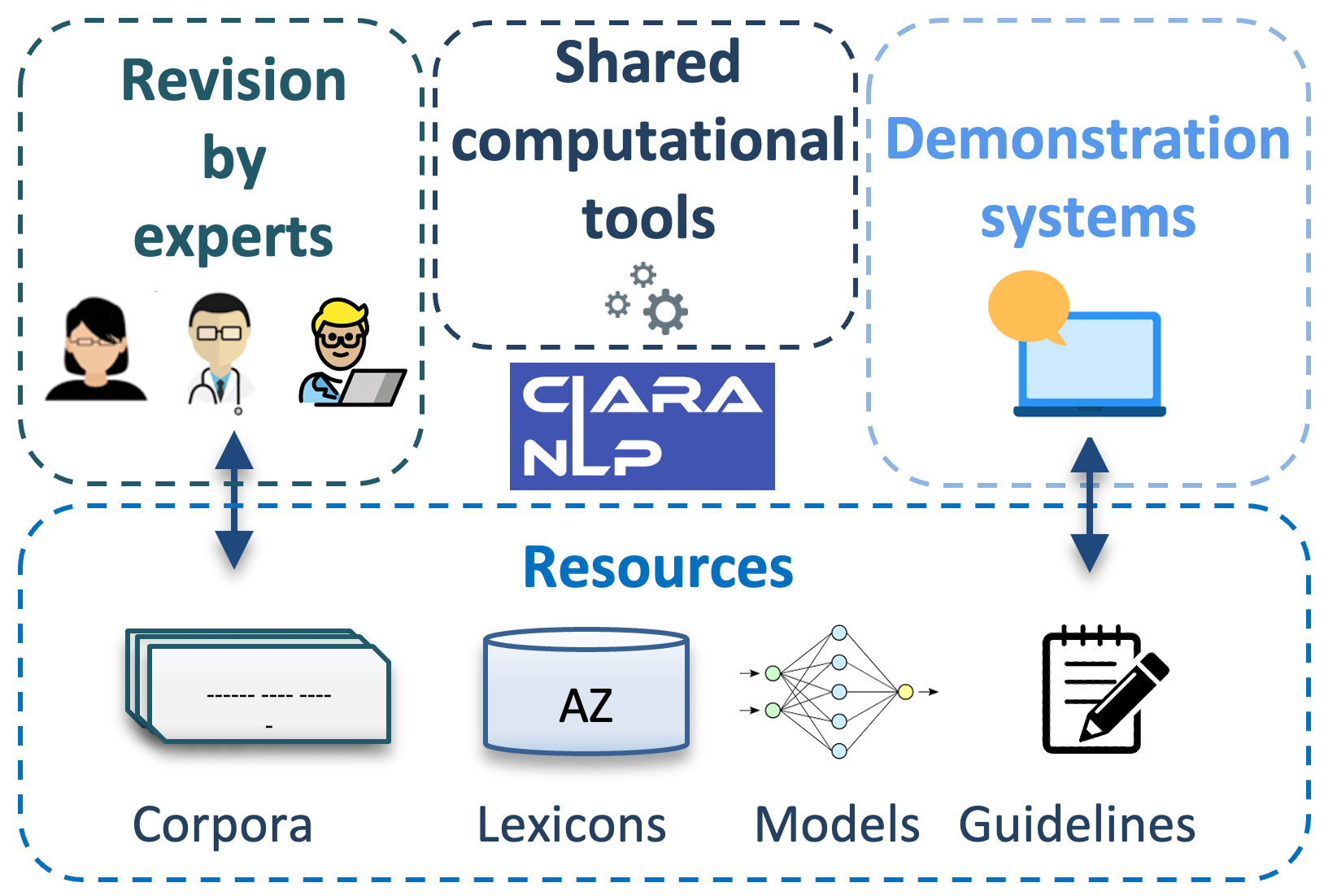}
 \caption{\label{fig:overview} Overview of CLARA-NLP. } 
\end{figure}
Figure \ref{fig:overview} is an overview of the produced resources. We rely on shared computational tools such as annotation and transcription software---BRAT \cite{stenetorp2012brat} and Transkribus \cite{kahle2017transkribus}---and a GPU. The language resources (corpora, lexicons, trained models or simplification guidelines) were developed in close collaboration with domain experts. These resources were released in public repositories\footnote{\url{https://clara-nlp.uned.es/}} or integrated into systems to simplify financial or medical texts or transcribe historical documents (and simplify them in the next stage).

\section{Corpora }
\label{results}
In the following sections, we describe the different domain-related corpora.
\label{corpora}

\subsection{Digital Humanities}

CLARA-DM is a historical corpus from an image-based digitised collection of newspapers called \textit{Diario de Madrid} (DM) from the Spanish press between the 18th and 19th centuries. DM is freely available at the Spanish National Library (BNE). In this work, we used Transkribus for the transcription. Once the sub-corpus of historical newspapers was manually transcribed, we used it to seed the Transkribus tool and started developing a DL model for the transcription task. According to the Transkribus guidelines, it is possible to begin training the model with at least 75 manually-transcribed pages. The tool also provides different export formats to further management of the transcribed corpus \cite{salidoetal2023}.   

\begin{figure}[htb!]
 \includegraphics[width=7.5cm,clip]{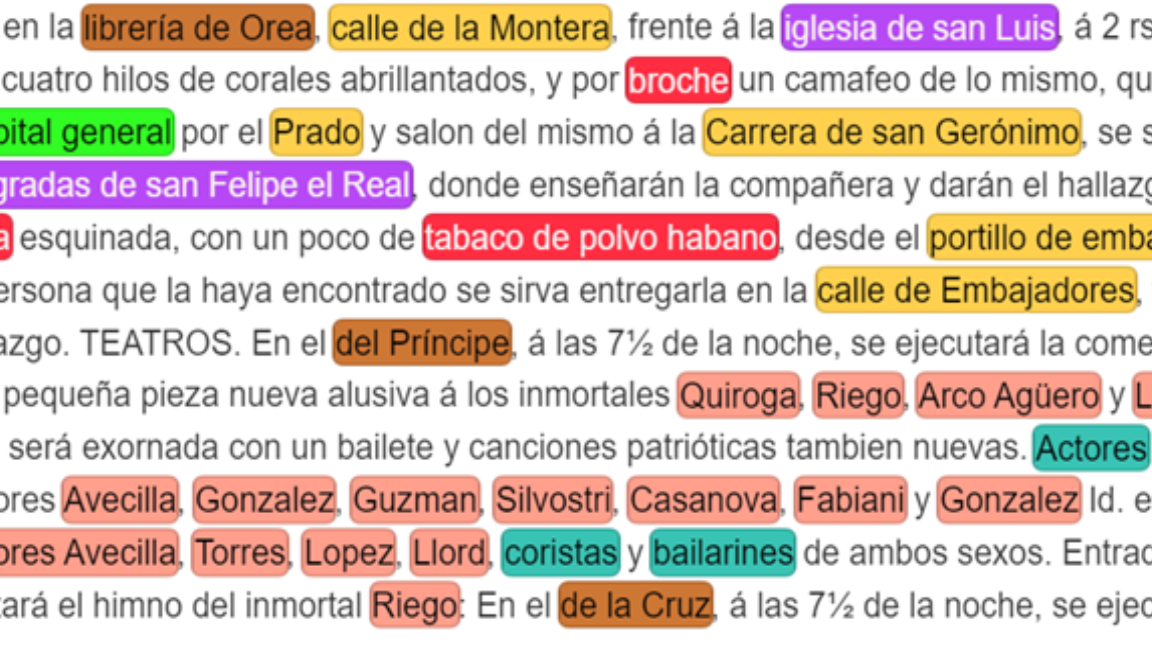}
 \caption{\label{fig:DH-corpus-sample} Hard-copy of a manually annotated text in the Tagtog tool; image from \cite{salidoetal2023}.} 
\end{figure}

The DM original collection has 13 479 newspapers (59 424 pages), and the CLARA-DM corpus currently has 37 manually-transcribed newspapers (201 pages), 143 automatically-transcribed newspapers (657 pages), and ten manually-annotated newspapers (53 pages; 24 843 tokens). 
The first sub-corpus contains five newspapers with 28 pages, 928 sentences and 15 145 tokens, annotated by four annotators in a blind annotation process, following the guidelines (\S\ref{guidelines}). The transcription, annotation, and NER tasks are the first stages before the simplification is conducted in a future step. Figure \ref{fig:DH-corpus-sample} is a sample of a manually annotated text.

\subsection{Finance}

FinT-esp \cite{fint-esp2020} is the corpus of annual reports collected by the UAM team in a previous project. The work presented focused on expanding the corpus by compiling parallel bilingual texts. Some IBEX companies have published their reports in Spanish and English in recent years. This made it possible to have a bilingual dataset that could be exploited in bilingual terminology extraction. The whole compilation process has been described in \cite{torterolo2023-CL}.

\subsection{Medicine}

In the medical domain, the corpus \cite{campillos2022building} is a comparable collection (with technical and non-specialists versions) of 24 298 pairs of professional and simplified texts ($>$96 million tokens).\footnote{\scriptsize \url{https://digital.csic.es/handle/10261/269887}} 
We used summaries of product characteristics and patient leaflets from the Spanish Medical Drug Information Centre (CIMA),\footnote{\url{https://cima.aemps.es}} excerpts from clinical trial announcements in the European clinical trials registry (EudraCT)\footnote{\url{https://www.clinicaltrialsregister.eu/}} and cancer-related information summaries from the National Cancer Institute.\footnote{\url{https://www.cancer.gov/espanol}} The corpus may be used to train medical simplification models and can be downloaded for research.

In addition, we collected 5000 professional/simplified sentence pairs for experiments at the sentence level. A subset of 3800 pairs were extracted from the comparable corpus, aligned using sentence-embeddings \cite{reimers-2019-sentence-bert} and manually revised. Another subset of 1200 sentences was simplified manually  \cite{campillos2024sentences}, following our simplification guidelines (\S\ref{guidelines}). We achieved high Inter-Annotator Agreement (IAA) for the 3800 sentences (average Cohen's $K$ = 0.839) and high scores in a human evaluation of a subset from the 1200 sentences (average of 4.7 on a 5-point Likert scale).


\section{Guidelines}
\label{guidelines}

In the following, we describe the needed guidelines for the corpus and model training.

\subsection{Digital Humanities}

In the Digital Humanities (DH) domain, applying NER models to historical documents poses several challenges \cite{chastang2021named}; for example, the transfer of knowledge to new domains or languages \cite{baptiste2021transferring,ehrmann2022extended}. We use the HIPE project approach \cite{ehrmann2020extended}, for which we have few annotated data.

The developed annotation guideline entails identifying the domain-based categories of labels (semantic categories) of interest for art historians, such as objects (e.g., lost-and-found items, for sale, and others). In this work, we use Tagtog for manual annotation\footnote{\url{https://tagtog.com} } (unfortunately, the web application is not currently available).

Based on the dialogue with the art historians and the documents they provided us with, a proposal for a taxonomy of entities was developed.  
The annotation guidelines were revised in several rounds, by analysing both the IAA and the performance of the models. 
Eleven new newspapers were annotated following the latest guidelines. 

Once all the occurrences in the sub-corpus of historical newspapers were manually annotated and tagged, the Tagtog tool provided different export formats to start developing the DL model for the NER task (\S\ref{models}).

\subsection{Finance}

In the financial domain, we developed three annotation guidelines: one for a discourse marker tagger, another for the term/keyword extractor (described in \S\ref{models}), and the guideline for the Spanish FinCausal dataset (\S\ref{shared-tasks}).

\subsection{Medicine}

In the CLARA-MeD sub-project, we created sentence simplification guidelines. In the first corpus version \cite{campillos2022building}, we adapted the criteria adopted by other teams for the medical domain \cite{grabar2018clear}. 
In the second stage \cite{campillos2024sentences}, we developed new guidelines for syntactic and lexical simplification and simplified 1200 sentences manually. Two simplified versions were obtained: one with syntactic simplification and another with both lexical and syntactic simplification. This more comprehensive version was used in experiments with DL models (\S\ref{models}).  

We followed the recommendations by experts in plain language \cite{cunha2022redactor} and former work on lexical simplification \cite{koptient2020fine,carbajo2023financial}, along with directives by organisations such as the International Plain Language Federation and the European Commission \cite{escribirclaridad}. 
Syntactic simplification operations involved changing nouns or adjectives to verbs, changing passive to active voice, removing appositive phrases, deleting unnecessary conjunctions, eliminating redundancies and co-reference/anaphora ambiguities, processing ambiguous negation, and splitting long sentences. Lexical simplification involved expanding acronyms/abbreviations, adding or deleting redundant lexemes, replacing technical terms with hypernyms, paraphrases/definitions or simpler synonyms, or translating English terms to Spanish.

\section{Lexicons}
\label{lexicon}
For the financial domain, we extracted 13 869 terms automatically from a collection of 315 annual reports from Spanish IBEX companies. This corpus was gathered in a previous project and contains more than 11 million tokens. 
The extracted terms were further reviewed by linguists, with the assistance of a financial expert. In the CLARA-FIN sub-project, we collected a bilingual corpus and applied a transformer-based automatic term extractor to extend our lexicon in Spanish and English. Our goal is to include approximately 20 000 terms and concepts from the financial domain that have been recently used by experts in real documents.

For the medical domain, we collected the SimpMedLexSp lexicon, which includes pairs of technical and simplified terms (e.g. \textit{cefalea}, `cephalalgia' $\leftrightarrow$ \textit{dolor de cabeza}, `headache'). The current version amounts to 14 013 pairs with gender, number and conjugated verb forms. To create this lexicon, we used several sources: EUGLOSS \cite{eugloss-glossary}, colloquial terms from \textit{Diccionario de términos médicos} \cite{dtm2011}, pairs of acronyms/abbreviations and full terms from the MedLexSp lexicon \cite{campillos2023medlexsp} and technical and simplified terms extracted from the CLARA-MeD corpus using regular expressions and paraphrase patterns. SimpMedLexSp is freely available for research or educational purposes.

\section{Trained Models}
\label{models}

Different domain-dependent DL models are described in the following subsections.
\subsection{Digital Humanities}

We developed two DL Models for this domain. 
First, an automatic transcription model (Spanish print XVIII-XIX) was already published in January 2023 for free use at the Transkribus tool website.\footnote{\scriptsize\url{https://readcoop.eu/model/spanish-print-xviii-xix}} Our starting point was selecting an existing DL model, namely the Spanish Golden Age Theatre Manuscripts (Spelling Modernization) 1.0 \cite{cuellar2021}. 

The second model is aimed at a NER task on the CLARA-DM corpus. The need for a new NER model is discussed in \cite{salidoetal2023}. In the first step, we experimented with monolingual RoBERTa models \cite{liu2019roberta} and multilingual XLM-RoBERTa~\footnote{\scriptsize\url{https://huggingface.co/xlm-roberta-base}} \cite{conneau-etal-2020-unsupervised}. Also, we used models that were trained for a general set of NER tags: a monolingual model for Spanish (RoBERTa-BNE-NER-CAPITEL);\footnote{\scriptsize\url{https://huggingface.co/PlanTL-GOB-ES/roberta-base-bne-capitel-ner}} and two multilingual ones: 
XLM-RoBERTa-NER-Spanish and XLM-RoBERTa-NER-HRL (trained in 10 languages and with more extensive resources).

In the second set of experiments, we used the manually annotated sub-corpus of 5 newspapers after a pre-processing phase to delimit the sentences (using the spaCy library).\footnote{\url{https://spacy.io}} 
We obtained a dataset of 928 sentences, comprising 15 145 tokens. The results of the experiments could have been better by improving the quality of the annotation guidelines and Inter-annotator Agreement (IAA). We also conducted an experiment applying a different adjudication method (Majority Vote versus Best Annotators method).

In summary, from a collection of newspapers in PDF format, we obtained a transcription model with an accuracy of 99\% and a NER model for the transcribed newspapers with an accuracy of more than 75\%. 

We also performed some other experiments for sentence simplification but did not use the CLARA-DM corpus. More details are explained in \cite{menta2022controllable}.

\subsection{Finance}

For the financial domain, we explored automatic concept extraction by employing transformer-based models \cite{porta2024extraction}. We focused mainly on lexical simplification in Spanish \cite{carbajo2023financial}. 

At the syntactic level, we developed an automatic discourse marker (DM) tagger for Spanish, which achieved significant agreement rates among human annotators and a very high F1-score using a transformer model \cite{toro2022discourse}.

\subsection{Medicine}

For medical texts, we used the 5000 parallel sentences from the CLARA-MeD corpus (\S\ref{corpora}) to train several models \cite{campillosetal2024}. We applied transformer-based architectures: multilingual BART \cite{liu-etal-2020-multilingual-denoising}, mT5 \cite{xue-etal-2021-mt5}, NASES \cite{ahuir2021nasca} or Pegasus XSUM \cite{zhang2019pegasus}; and prompt-based learning using BERTIN Alpaca \cite{BERTIN-prompt-learning}. In a quantitative evaluation using BLEU \cite{papineni2002bleu}, ROUGE \cite{rouge2004package} and SARI \cite{xu2016optimizing}, the mBART model outperformed the other approaches, including the method based only on lexical simplification and/or paraphrasing (using the SimpMedLexSp lexicon). 

\begin{figure*}[htbp!]
\begin{center}
\begin{tabular} {|c|c|}
\hline
 \includegraphics[width=7.5cm,clip]{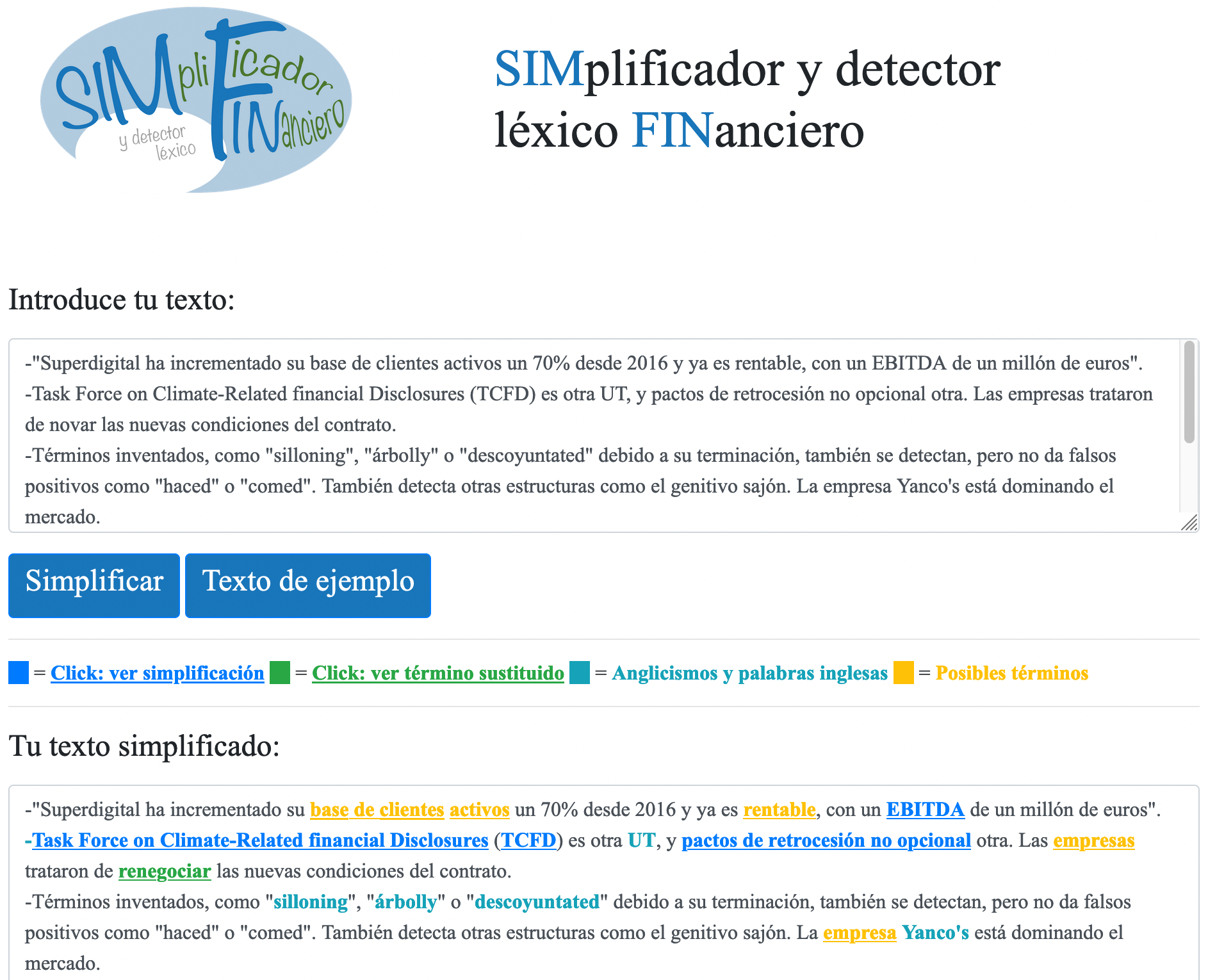} & \includegraphics[width=7.5cm,clip]{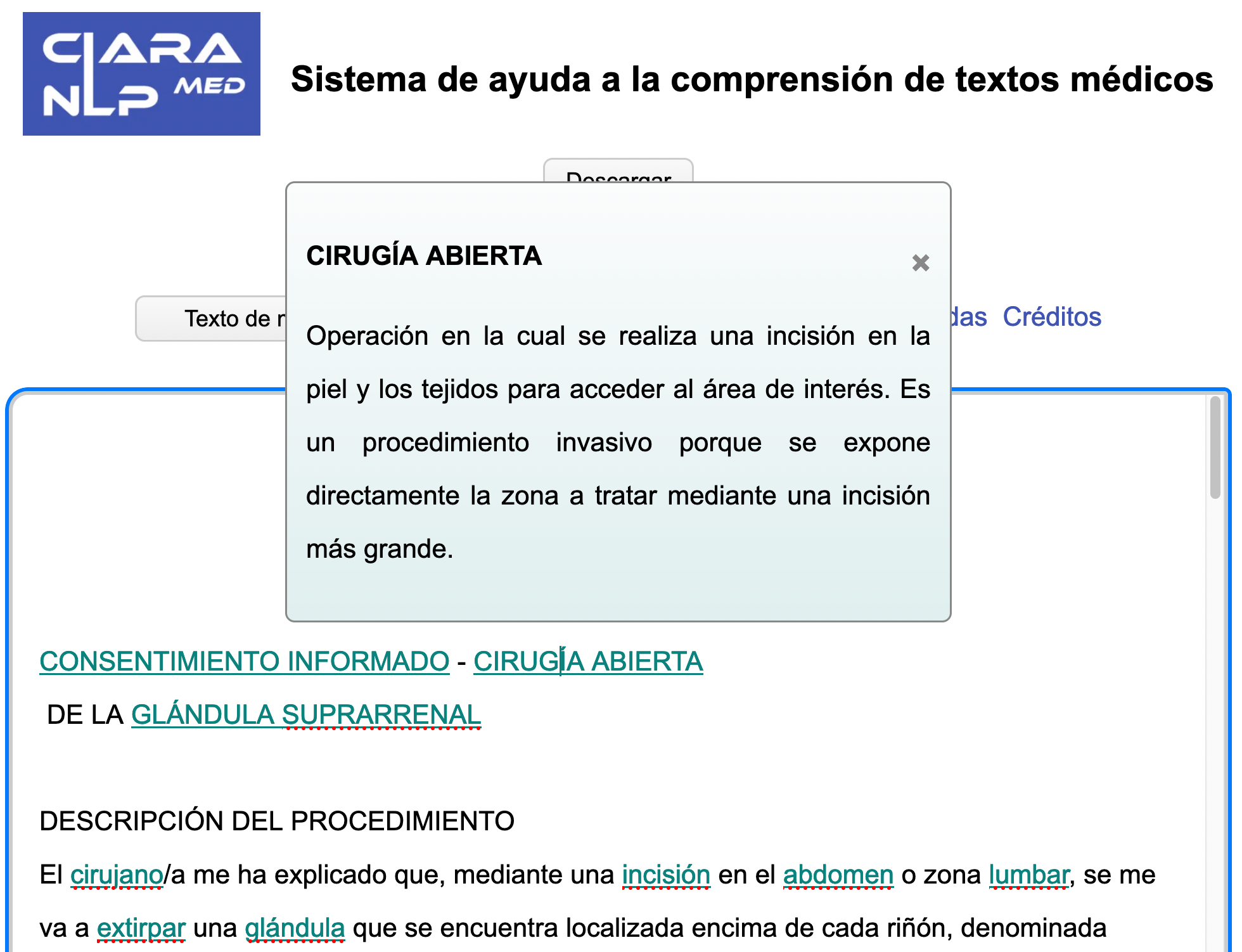} \\
\hline
\end{tabular}
\end{center}
\vspace{-1em}
\caption{\label{fig:demos1}Simplification tools for the financial domain (left) and for the medical domain (right).}
\end{figure*}

We also conducted a qualitative evaluation in which 11 human subjects assessed 500 simplified sentences according to grammar/fluency, meaning adequacy and overall simplification. This evaluation showed that the quantitative metrics did not correlate with human judgements. Indeed, the models with the highest quantitative metrics produced hallucinations. The prompt-based approach using BERTIN Alpaca (trained with SimpMedLexSp) obtained balanced scores in all evaluated aspects. Our findings indicate that combining a lexical resource and DL methods achieved the highest quantitative and qualitative scores.

Note that, in the CLARA-MeD sub-project, we also trained several NER models for information extraction aimed at medical professionals. Since these resources are beyond the scope of the simplification task, we refer to \cite{campillos2024bioinf} for more details.

\section{Simplification Systems and Participation in Shared-tasks}
\label{shared-tasks}

Figure \ref{fig:demos1} show two screenshots of the simplification systems. For the medical domain, the CLARA-MeD simplification system 
allows users to input a medical text (e.g. a drug leaflet or a clinical trial announcement) and have medical terms highlighted along with an explanation or a simpler synonym. The tool has been evaluated with patient-information documents, consent forms and clinical trial announcements \cite{campillosMIE2024}. Still, an end-user evaluation is a pending task for future work. In addition to this tool, we distribute companion resources (e.g. extracted n-grams) and Python scripts for word frequency filtering, readability analyses and parallel (professional/non-specialists) sentence alignment (available at the companion GitHub repository).\footnote{\scriptsize\url{https://github.com/lcampillos/CLARA-MeD}} %

SimpFin is a financial term simplification tool designed to make complex financial terminology more accessible for non-experts. 

P. Rayson and M. El-Haj, from Lancaster University, were the main organisers of the Financial Narrative Processing series.\footnote{\url{https://wp.lancs.ac.uk/cfie/}} The UAM team participated in the creation of the Spanish datasets for the following competitions: 
\begin{itemize}
    \item FNP2022 at LREC, Marseille, France: Financial Narrative Summarisation (FNS), an initiative for summarising financial annual reports from the UK, Greece, and Spain \cite{el-haj-etal-2022-financial}; and Financial Document Structure Extraction (FinTOC), which focused on extracting and hierarchically organising the structure of financial texts, fostering progress in table-of-contents extraction technologies \cite{kang2022financial}.
    \item FNP2023 at IEEE Big Data conference, Sorrento, Italy: we again annotated the FNS Spanish datasets \cite{zavitsanos2023financial}. This time, we organised the Financial Document Causality Detection Shared Task (FinCausal 2023) in English and Spanish. The primary objective of this task is to identify whether an object, event or sequence of events can be considered the cause of a preceding event, i.e., the effect \cite{moreno2023financial}. 
\end{itemize}

As participants, we submitted our system to the Financial Targeted Sentiment Analysis in Spanish (FinancES) at IberLEF 2023 \cite{porta2023lli}. This paper presents a T5-based system developed for the shared task. 
It included noise and data augmentation experiments, using corrected datasets and ChatGPT for data improvement. The paper reports on the system’s performance across tasks, detailing the impact of noise, data augmentation, and hallucinations on model accuracy. We ranked \(2^{nd}\) in the first task (Target detection) and \(1^{st}\) in the second task (Multi-dimension sentiment classification).

\section{Discussion and Conclusions}

In this work we focused on creating simplified versions of texts from different domains, such as financial reports, 18th- and 19th-century newspapers, and medical texts. We employed lexicons and deep learning (DL) methods to produce datasets, resources and simplified texts. However, we found that these approaches did not always capture all the relevant information or even produced hallucinations and factual errors. 

For example, some explanations of events were provided in financial reports, but the causes were left out. Moreover, the causes of a problem or an event were sometimes mentioned, not the consequences. In simpler terms, it is challenging to identify omitted or over-understood information, which can affect the accuracy of simplification systems and other generative artificial intelligence (AI) models like translation, automatic summarisation, or question answering (QA).

When it comes to medical simplification, models achieving the highest quantitative scores (e.g., mBART) produced hallucinations or created non-existent medical terms (e.g. \textit{*tendinios} instead of \textit{tendones}, `tendons'). These severe errors might spread fake data or misinformation and are to be avoided in a sensitive domain such as Medicine. Nevertheless, using a domain lexicon increased the performance of the trained DL models. 

We have recently started exploring the use of generative AI systems, such as ChatGPT \cite{chatgpt} or Gemini \cite{gemini}, with our corpora. In the CLARA-Fin sub-project, however, we found that using ChatGPT to augment the training dataset did not yield significant results in the system used in the FinancES challenge. Similarly, in FinCausal2023, two teams from India also used ChatGPT with prompting to locate cause and effect segments, but they did not perform well in the competition. 

In the CLARA-MeD sub-project, we used ChatGPT vs. 3.5 to explain or simplify 1369 medical terms that were missing in our lexicons; nonetheless, definitions were validated by a medical doctor or domain documentalists/archivists. The current technology requires human supervision, particularly in specialised domains. 

In the CLARA-DH sub-project, we use ChatGPT to simplify very long sentences with old words and spelling errors. The text generated by ChatGPT is manually corrected by annotators and is displayed in the application interface of an annotation tool along with the original sentence. The annotators will generate a new corpus of CLARA-DH of sentence pairs. These texts will be used as seeds for developing a sentence simplification model of the CLARA-DM corpus, given that it has more than 600 sentence pairs to date.

As part of our future work, we aim to explore generative AI using the Retrieval-Augmented Generation (RAG) technology approach introduced by \cite{gao2023retrieval}. This approach combines natural language generation (NLG) with information retrieval (IR) to improve the quality of answers generated by large language models (LLMs). The most important contribution of this new technique is that it corrects or avoids the hallucinations and knowledge cutoff problems, typical of LLM-based systems such as ChatGPT or Gemini. 

In conclusion, the experience gained in our project has led us to distribute language resources (corpora, lexicons, guidelines and models) that are the cornerstone for quality outcomes in simplification tasks. These resources have been created or validated in direct collaboration with historians, financial specialists or professionals from the medical domain. This validation by experts and the fine inter-annotator-agreement achieved ensures that the contents are valid for experimental tasks. Even so, the size and coverage of the current resources need to be enlarged in future versions. Concerning DL approaches and generative AI, our primary outcomes suggest that language resources are required for optimal results, boosting the performance of DL models alone. Still, human evaluation and experts' supervision are imperative to avoid hallucinations, factuality issues and misinformation.



\bibliographystyle{fullname}
\bibliography{claranlp2024}

\appendix

\end{document}